\documentclass[english]{article}
\usepackage[latin9]{luainputenc}
\usepackage{amsbsy}
\usepackage{amstext}
\usepackage{graphicx}

\makeatletter

\providecommand{\tabularnewline}{\\}
\newcommand{\lyxdot}{.}

\usepackage{microtype}
\usepackage{flushend}
\usepackage{ijcai17}\usepackage{times}\usepackage{helvet}\usepackage{courier}

\usepackage{amsbsy}
\usepackage{algorithmic}
\usepackage[english]{babel}

\makeatother

\usepackage{babel}
\begin{document}

\title{Deep Learning to Attend to Risk in ICU}

\author{Phuoc Nguyen, Truyen Tran, Svetha Venkatesh\\
Centre for Pattern Recognition and Data Analytics\\
Deakin University, Geelong, Australia\\
\{\textit{phuoc.nguyen, truyen.tran, svetha.venkatesh}\}\textit{@deakin.edu.au}}

\maketitle
\global\long\def\xb{\boldsymbol{x}}
\global\long\def\yb{\boldsymbol{y}}
\global\long\def\hb{\boldsymbol{h}}
\global\long\def\Xcal{\mathcal{X}}
\global\long\def\Ucal{\mathcal{U}}
\global\long\def\Vcal{\mathcal{V}}
\global\long\def\Real{\mathbb{R}}
\global\long\def\mat{\text{mat}}
\global\long\def\thetab{\boldsymbol{\theta}}
\global\long\def\Wb{\boldsymbol{W}}
\global\long\def\bb{\boldsymbol{b}}
\global\long\def\cb{\boldsymbol{c}}
\global\long\def\gb{\boldsymbol{g}}
\global\long\def\fb{\boldsymbol{f}}
\global\long\def\ob{\boldsymbol{o}}
\global\long\def\ib{\boldsymbol{i}}
\global\long\def\wb{\boldsymbol{w}}
\global\long\def\ab{\boldsymbol{a}}
\global\long\def\zb{\boldsymbol{z}}
\begin{abstract}
Modeling physiological time-series in ICU is of high clinical importance.
However, data collected within ICU are irregular in time and often
contain missing measurements. Since absence of a measure would signify
its lack of importance, the missingness is indeed informative and
might reflect the decision making by the clinician. Here we propose
a deep learning architecture that can effectively handle these challenges
for predicting ICU mortality outcomes. The model is based on Long
Short-Term Memory, and has layered attention mechanisms. At the sensing
layer, the model decides whether to observe and incorporate parts
of the current measurements. At the reasoning layer, evidences across
time steps are weighted and combined. The model is evaluated on the
PhysioNet 2012 dataset showing competitive and interpretable results.

\end{abstract}

\section{Introduction}

Multivariate physiological time-series are a critical component in
monitoring the critical state of the patient admitted to ICU \cite{ghassemi2015state}.
Characterizing this data type must take into account the fact that
\emph{data are irregularly sampled}. That is, biomarkers (Cholesterol,
Glucose, Heart rate, etc.) are measured and recorded only then the
attending doctors decide to do so, e.g., a particular measurement
is made to find out a critical condition about the patient at a given
time. Other measurements may be omitted if they offer no new knowledge,
are expensive and invasive, or if the patient is stable with respect
to these physiological parameters. Other reasons for missing data
could be due to technical errors, that is, the measured signals are
unreliable or interrupted in the urgent and intensive conditions in
ICU. 

\begin{figure}[h]
\begin{centering}
\includegraphics[width=0.85\columnwidth]{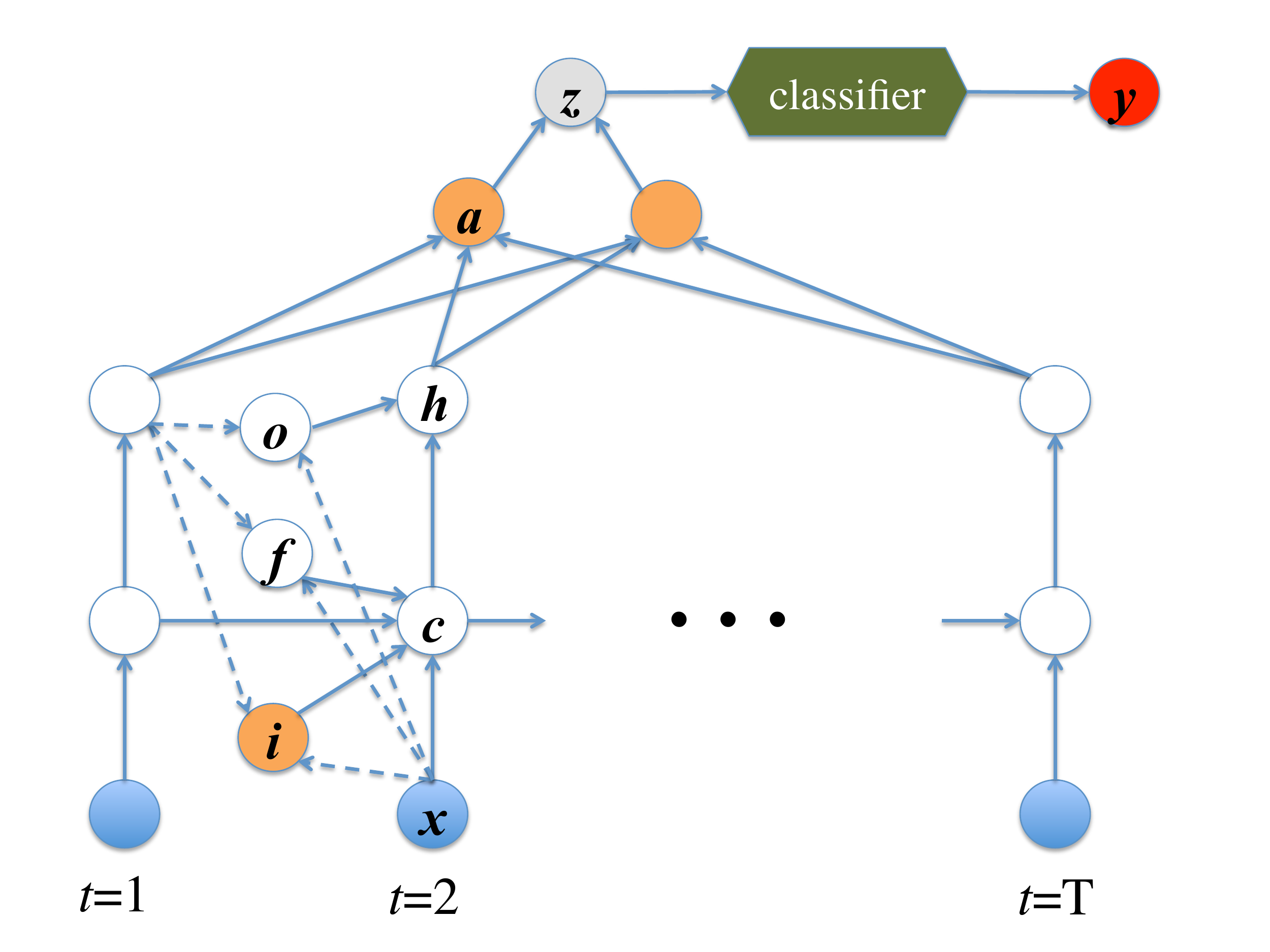}
\par\end{centering}
\caption{Recurrent neural net (LSTM) equipped with multiple attentions for
outcomes prediction in ICU. Each time step $t$, time-series within
the interval are processed (with missing data imputed and statistics
collected) into an input vector $\protect\xb_{t}$. The input gate
$\protect\ib_{t}$ decides what to reads from the input. The forget
gate $\protect\fb_{t}$ controls the refreshing rate of the memory
$\protect\cb_{t}$. The states $\protect\hb_{1:T}$ are pooled using
multiple read\textendash heads (attentions) $\protect\ab_{1:R}$ to
produce the feature vector $\protect\zb$, which is used by the differentiable
classifier to predict the outcomes $\protect\yb$. The attention nodes
are in orange. Best viewed in color.\label{fig:model}}
\end{figure}

Irregular timing causes great difficulties in statistical analysis
as the time gaps cannot easily characterized or predicted. In this
paper we consider predicting ICU mortality from physiological time-series.
Unfortunately, most existing time-series methodologies assume equally
spaced data \cite{erdogan2004statistical}. To this end, we propose
a method that partly deals with the difficulties due to irregular
sampling. The main idea is to ``attend'' to important signals/biomarkers
and ignore others. There are two level of attentions. At the signal
level, only informative signals at given time are kept. At the illness
state level, temporal state progression is considered, and only time
at which states are most critical for prediction will be kept. The
mechanism for weighting the temporal importance is called ``attention''.
The attention mechanism might capture the implicit human decision
making.

To realize the attention ideas, we derive a deep learning architecture
based on a recurrent neural network known as Long Short-Term Memory
(LSTM) \cite{hochreiter1997long}. The LSTM offers an attention mechanism
at the signal level through input gating. On top of the LSTM, we impose
several attention ``heads'' that read the illness states over time,
and decide on the importance of each time interval. The resulting
model is flexible and interpretable. Fig.~\ref{fig:model} illustrates
the model. Evaluation of the PhysioNet 2012 dataset demonstrates the
desirable characteristics.

\section{Methods\label{sec:Methods}}

We present our deep neural net for reading ICU time-series and predicting
mortality. The model consists of four components: \emph{data preprocessing},
\emph{LSTM,} \emph{reading heads} and \emph{classifier}. The last
three components are graphically depicted in Fig.~\ref{fig:model}.

\subsection{Data preprocessing}

We assume that each patient has multiple variable-length time-series
sampled at arbitrary times. As the data is highly irregular, it might
be more useful to partly transform the data so that some regularities
can emerge. First for robustness, outliers are handled by truncating
all measures into the range of $[0.01,0.99]$ percentiles. Then we
divide the entire time-series into intervals of equal length (e.g.,
3 hours). If a measure is missing in an interval, it is imputed by
its mean value across time for the patient. If the patient does not
have this biomarker measured, the mean value is taken from the entire
dataset. For each interval we collect simple statistics on each physiological
measure, including \{\emph{min, max, mean, median} and \emph{standard
deviation}\}. Finally, for each patient, we have a sequence of vectors,
where each vector is the set of statistics for its interval.

\subsection{Long Short-Term Memory}

A Long Short-Term Memory (LSTM) \cite{hochreiter1997long} is a recurrent
neural network. Let $\xb_{t}$ denotes the input vector at time $t$.
LSTM maintains a memory cell $\cb_{t}$ and a state vector $\hb_{t}$
over time. In our model, this state vector is considered representing
a patient's illness state from begin to the current time $t$. Let
$\tilde{\cb_{t}}$ be a candidate new memory update, which is a function
of the previous state $\hb_{t-1}$ and the current input gate $\ib_{t}$.
The memory is updated as follows:
\begin{equation}
\cb_{t}=\fb_{t}\ast\cb_{t-1}+\ib_{t}\ast\tilde{\cb_{t}}\label{eq:memory-update}
\end{equation}
where $\ast$ is point-wise multiplication, $\ib_{t}\in(\boldsymbol{0},\boldsymbol{1})$
is the input gate to control what to read from raw data, and $\fb_{t}\in(\boldsymbol{0},\boldsymbol{1})$
is the forget gate to control the refreshing rate of the memory. The
two gates are function of $\hb_{t-1}$ and $\ib_{t}$. 

This equation is crucial for handling irregular information. When
the new input is uninformative, the input gate $\ib_{t}$ can learn
to turn off (i.e., $\ib_{t}\rightarrow\boldsymbol{0}$) and the forget
gate can learn to turn on (i.e., $\fb_{t}\rightarrow\boldsymbol{1}$),
and thus $\cb_{t}\rightarrow\cb_{t-1}$. In other words, the memory
is maintained. On the other hand, when the new input is highly informative,
the old memory can be safely forgotten (i.e., $\ib_{t}\rightarrow\boldsymbol{1}$,
$\fb_{t}\rightarrow\boldsymbol{0}$ and $\cb_{t}\rightarrow\tilde{\cb_{t}}$).

Finally, the state vector is computed as
\begin{equation}
\hb_{t}=\ob_{t}\ast\text{tanh}(\cb_{t})\label{eq:lstm-state}
\end{equation}
where $\ob_{t}\in(\boldsymbol{0},\boldsymbol{1})$, which is a function
of $\hb_{t-1}$ and $\ib_{t}$. 

\subsubsection{Bidirectional LSTM}

The LSTM is directional from the past to the future. Thus when a state
is estimated, it cannot be re-estimated on the face of new evidences.
A solution is to use bidirectional LSTM, that is, we maintain two
LSTMs, one from the past to the future, the other in the reverse direction.
The joint state $\hat{\hb}_{t}=\left[\overrightarrow{\hb}_{t},\overleftarrow{\hb}_{t}\right]$
is likely to be more informative than each of the component.

\subsection{Reading heads}

For each patient, the LSTMs produce a sequence of state vectors $\hat{\hb}_{1:T}$.
It is like a memory bank of $T$ slots from which a read head can
operate to generate sequence-level outputs. Since $T$ is usually
variable, we need to aggregate all the states into a fixed-size vector.
A number of \emph{soft} reading heads are therefore employed:
\begin{equation}
\bar{\hb}_{r}=\sum_{t=1}^{T}a_{rt}\hat{\hb}_{t}\label{eq:reading}
\end{equation}
where $r=1,2,..,R$ is the index of the reading head, $a_{rt}\ge0$
and $\sum_{t=1}^{T}a_{rt}=1$. Here $a_{rt}$ is known as the \emph{attention
mechanism}, and is parameterized as a neural network:
\begin{equation}
a_{rt}=\frac{\exp\left(\text{nnet}(\hat{\hb}_{t})\right)}{\sum_{j=1}^{T}\exp\left(\text{nnet}(\hat{\hb}_{j})\right)}\label{eq:attention}
\end{equation}
Finally, readings are max-pooled as: $\zb=\max_{r}\left\{ \bar{\hb}_{r}\right\} $.

\subsection{Classifier}

Given the fixed-size vector $\zb$, any differentiable classifier
can be placed on top to predict the future. For example, to predict
ICU mortality, a simple logistic regression can be used: $P\left(y=1\mid\xb_{1:T}\right)=\text{sigmoid}\left(\wb^{\top}\zb+w_{0}\right)$
for regression parameters $(w_{0},\wb)$, and positive outcome $y=1$.
Finally, the entire system is learnt by minimizing the log-loss: $L=-\log P\left(y\mid\xb_{1:T}\right)$.
Since the system is end-to-end differentiable, automatic differentiation
and gradient descent methods can be employed.

\section{Experiments}

\subsection{Dataset and Setting}

We use data from the PhysioNet Challenge 2012 \cite{silva2012predicting}.
There are 4,000 patients of age 16 or over (mean: 64.5, std: 17.1),
56.1\% are males. Of 41 measure types, five are static (age, gender,
height, ICU type, and initial weight), and the other 36 are time-series.
The recording time is 48 hours max. There are 4 ICU types: medical
(35.8\%), surgical (28.4\%), cardiac surgery recovery (21.1\%), and
coronary (21.1\%). The overall mortality rate is 18\%. 

The models are implemented using Knet.jl. For the experiments reported
below, time intervals are 3 hours long, resulting in 16 intervals
per patient at most. At each interval, 185 statistics are extracted
as input features. The memory cell (hence the state and output vector)
size is 32. The state vector of the memory cell can be thought of
as representing a patient's illness state. At time $0$, this state
is set to zero vector. At each following time step, it is changed
according to the response of the input signals to the trained model.
The progression of this state vector is different between a positive
and a negative case. Two read heads are used to generate the output
features, which is then fed to a simple logistic regression to estimate
the probability of death. Dropout is utilized at both the input features
(due to high level of redundancy in the extracted statistics) and
the output features. Prediction performance is evaluated using 5-fold
cross-validation.

\subsection{Results}

\begin{figure*}
\centering{}%
\begin{tabular}{cc}
\includegraphics[width=0.45\textwidth]{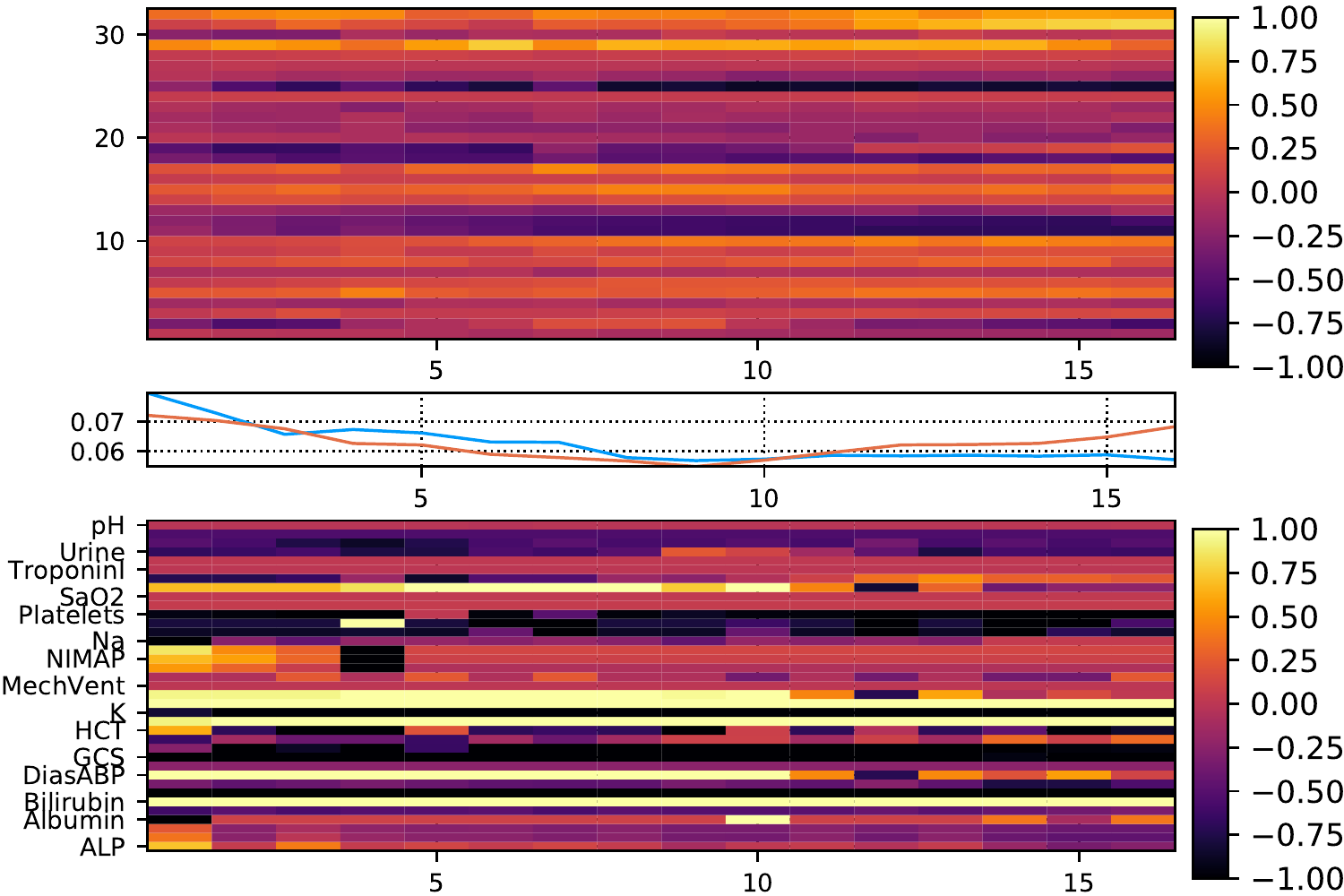}\quad{} & \quad{}\includegraphics[width=0.45\textwidth]{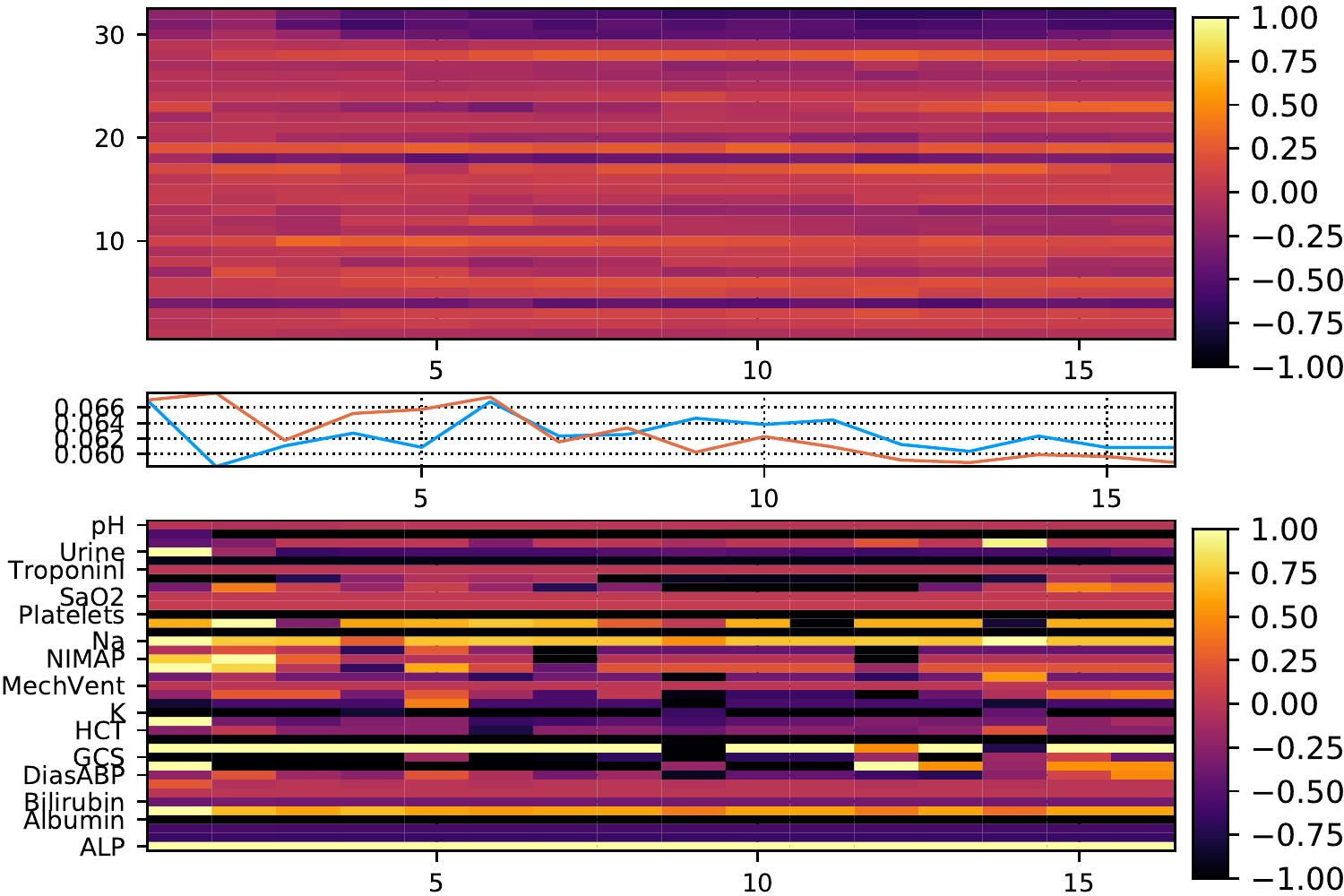}\tabularnewline
(a) $P=0.71$ & (b) $P=0.68$\tabularnewline
\end{tabular}\caption{Temporal dynamics: patient's illness states represented by the model
using $\protect\hb_{1:T}$ (top), attention probabilities $\protect\ab_{1:T}$
(middle) and 36 normalized measures $\protect\xb_{1:T}$ (bottom)
of two positive samples (mortality risks 0.71 and 0.68). \label{fig:Temporal-dynamics:-illness}}
\end{figure*}

Fig\@.~\ref{fig:Temporal-dynamics:-illness} depicts the temporal
dynamics of the model in time, in three ways: (a) progression of illness
states, (b) attention probabilities when estimating the risk of death,
and (c) actual mean measurements at each time interval. The 36 normalized
mean measures are shown below the attention probabilities. The attention
probabilities indicate when the information is more informative. It
suggests a simple alerting method when the attention probability are
beyond a certain threshold. This adds the third dimension for state
monitoring in addition to other two dimensions: the raw biomarker
readings and outcome risk probability.

Table~\ref{tab:all-methods} presents the Area Under the ROC Curve
(AUC) for competing methods. The baselines are (a) simple logistic
regression trained on statistics collected over the entire time-series;
(b) several simple imputation methods from the most recent work \cite{che2016recurrent},
and (c) LSTM without attention, where states are pooled by averaging.
It could be seen that (i) modeling the temporal dynamic using LSTM
is better than without, (ii) attention improves the prediction accuracy,
and (iii) bidirectional LSTM offers a marginal gain over LSTM when
attention is applied. 

\begin{table}
\begin{centering}
\begin{tabular}{ll}
\hline 
\emph{Method} & \emph{AUC}\tabularnewline
\hline 
48-hr stats + LR & 0.791\tabularnewline
LSTM-mean{*} & 0.803\tabularnewline
GRU-mean{*} & 0.820\tabularnewline
GRU-forward{*} & 0.816\tabularnewline
GRU-simple{*}  & 0.816\tabularnewline
\hline 
LSTM (mean pooling) & 0.825\tabularnewline
LSTM (with attention) & 0.833\tabularnewline
\textbf{BiLSTM (with attention)} & \textbf{0.839}\tabularnewline
\hline 
\end{tabular}
\par\end{centering}
\caption{AUC measures of different methods on PhysioNet 2012. LR = logistic
regression. BiLSTM = bidirectional LSTM. Time-intervals are 3 hours
long. ({*}) Results are from {[}Che \emph{et al}., 2016{]}. GRU (Gated
Recurrent Unit) is a recent alternative to LSTM. Mean, forward, simple
are imputation methods introduced in the cited paper. \label{tab:all-methods} }
\end{table}

\section{Related Work}

Irregular physiological time-series have attracted a fair amount of
attention in recent years, probably due to public availability of
large datasets such as MIMIC II/III \cite{caballero2015dynamically,che2016recurrent,durichen2015multitask,ghassemi2015multivariate,lasko2013computational,li2015classification,li2015physiological,lipton2016modeling,liu2016learning,liu2013modeling,schulam2016integrative,razavian2016multi}.
The most popular strategy to deal with missing data is imputation
based on interpolation \cite{eckner2012framework}. An alternative
method has also been suggested, in that the time gaps are part of
the models \cite{nguyen2016deepr,pham2017predicting}.

Neural nets in general and recurrent neural nets in particular have
long been applied for time-series data (e.g., \cite{tresp1998solution}).
The modern surge in deep learning has resulted in a new wave of more
powerful nets such as deep denoising autoencoder \cite{lasko2013computational}
and LSTM/GRU \cite{che2016recurrent,esteban2016predicting,lipton2016modeling}.
Attention has been recently suggested as a mechanism to boost interpretability
of RNNs \cite{choi2016retain}. The main difference is that the attention
in \cite{choi2016retain} is used to select the original data for
classification, where our attention is to select the illness state.

\section{Discussion}

We have proposed to use attention as a mechanism to mitigate the effect
of missing data resulted from irregular sampling in time-series. There
are two attention levels, one at the sensing layer to select informative
measurements, and the other at the reasoning layer to select the informative
period. The idea is realized using Long Short-Term Memory (LSTM) equipped
with multiple reading heads, which generate features for the classifier.
Experiments on ICU mortality prediction demonstrate that the models
are accurate and interpretable. It suggests that alert can be generated
in real-time if the new measurements are informative (based on the
attention probability) or the mortality risk is sufficiently high.

Future work includes more sophisticated imputation methods, such as
those in \cite{che2016recurrent}, handling multi-resolutions, and
explicitly incorporating data quality and uncertainty into reasoning.

\section*{Acknowledgements}

This work is partially supported by the Telstra-Deakin Centre of Excellence
in Big Data and Machine Learning.

\bibliographystyle{named}

\begin{thebibliography}{}

\bibitem[\protect\citeauthoryear{Caballero~Barajas and
  Akella}{2015}]{caballero2015dynamically}
Karla~L Caballero~Barajas and Ram Akella.
\newblock Dynamically modeling patient's health state from electronic medical
  records: A time series approach.
\newblock In {\em Proceedings of the 21th ACM SIGKDD International Conference
  on Knowledge Discovery and Data Mining}, pages 69--78. ACM, 2015.

\bibitem[\protect\citeauthoryear{Che \bgroup \em et al.\egroup
  }{2016}]{che2016recurrent}
Zhengping Che, Sanjay Purushotham, Kyunghyun Cho, David Sontag, and Yan Liu.
\newblock Recurrent neural networks for multivariate time series with missing
  values.
\newblock {\em arXiv preprint arXiv:1606.01865}, 2016.

\bibitem[\protect\citeauthoryear{Choi \bgroup \em et al.\egroup
  }{2016}]{choi2016retain}
Edward Choi, Mohammad~Taha Bahadori, Jimeng Sun, Joshua Kulas, Andy Schuetz,
  and Walter Stewart.
\newblock {RETAIN: An Interpretable Predictive Model for Healthcare using
  Reverse Time Attention Mechanism}.
\newblock In {\em Advances in Neural Information Processing Systems}, pages
  3504--3512, 2016.

\bibitem[\protect\citeauthoryear{D{\"u}richen \bgroup \em et al.\egroup
  }{2015}]{durichen2015multitask}
Robert D{\"u}richen, Marco~AF Pimentel, Lei Clifton, Achim Schweikard, and
  David~A Clifton.
\newblock {Multitask Gaussian processes for multivariate physiological
  time-series analysis}.
\newblock {\em IEEE Transactions on Biomedical Engineering}, 62(1):314--322,
  2015.

\bibitem[\protect\citeauthoryear{Eckner}{2012}]{eckner2012framework}
Andreas Eckner.
\newblock A framework for the analysis of unevenly spaced time series data,
  2012.

\bibitem[\protect\citeauthoryear{Erdogan \bgroup \em et al.\egroup
  }{2004}]{erdogan2004statistical}
Emre Erdogan, Sheng Ma, Alina Beygelzimer, and Irina Rish.
\newblock Statistical models for unequally spaced time series.
\newblock In {\em Proceedings of the Fifth SIAM International Conference on
  Data Mining}. SIAM, 2004.

\bibitem[\protect\citeauthoryear{Esteban \bgroup \em et al.\egroup
  }{2016}]{esteban2016predicting}
Crist{\'o}bal Esteban, Oliver Staeck, Stephan Baier, Yinchong Yang, and Volker
  Tresp.
\newblock Predicting clinical events by combining static and dynamic
  information using recurrent neural networks.
\newblock In {\em Healthcare Informatics (ICHI), 2016 IEEE International
  Conference on}, pages 93--101. IEEE, 2016.

\bibitem[\protect\citeauthoryear{Ghassemi \bgroup \em et al.\egroup
  }{2015a}]{ghassemi2015state}
Marzyeh Ghassemi, Leo~Anthony Celi, and David~J Stone.
\newblock State of the art review: the data revolution in critical care.
\newblock {\em Critical Care}, 19(1):1, 2015.

\bibitem[\protect\citeauthoryear{Ghassemi \bgroup \em et al.\egroup
  }{2015b}]{ghassemi2015multivariate}
Marzyeh Ghassemi, Marco~AF Pimentel, Tristan Naumann, Thomas Brennan, David~A
  Clifton, Peter Szolovits, and Mengling Feng.
\newblock {A Multivariate Timeseries Modeling Approach to Severity of Illness
  Assessment and Forecasting in ICU with Sparse, Heterogeneous Clinical Data}.
\newblock In {\em Twenty-Ninth AAAI Conference on Artificial Intelligence},
  2015.

\bibitem[\protect\citeauthoryear{Hochreiter and
  Schmidhuber}{1997}]{hochreiter1997long}
Sepp Hochreiter and J{\"u}rgen Schmidhuber.
\newblock Long short-term memory.
\newblock {\em Neural computation}, 9(8):1735--1780, 1997.

\bibitem[\protect\citeauthoryear{Lasko \bgroup \em et al.\egroup
  }{2013}]{lasko2013computational}
Thomas~A Lasko, Joshua~C Denny, and Mia~A Levy.
\newblock Computational phenotype discovery using unsupervised feature learning
  over noisy, sparse, and irregular clinical data.
\newblock {\em PloS one}, 8(6):e66341, 2013.

\bibitem[\protect\citeauthoryear{Li and Marlin}{2015}]{li2015classification}
Steven Cheng-Xian Li and Benjamin Marlin.
\newblock Classification of sparse and irregularly sampled time series with
  mixtures of expected gaussian kernels and random features.
\newblock In {\em 31st Conference on Uncertainty in Artificial Intelligence},
  2015.

\bibitem[\protect\citeauthoryear{Li-wei \bgroup \em et al.\egroup
  }{2015}]{li2015physiological}
H~Lehman Li-wei, Ryan~P Adams, Louis Mayaud, George~B Moody, Atul Malhotra,
  Roger~G Mark, and Shamim Nemati.
\newblock A physiological time series dynamics-based approach to patient
  monitoring and outcome prediction.
\newblock {\em IEEE journal of biomedical and health informatics},
  19(3):1068--1076, 2015.

\bibitem[\protect\citeauthoryear{Lipton \bgroup \em et al.\egroup
  }{2016}]{lipton2016modeling}
Zachary~C Lipton, David~C Kale, and Randall Wetzel.
\newblock Modeling missing data in clinical time series with rnns.
\newblock {\em Conference on Machine Learning in Healthcare (MLHC)}, 2016.

\bibitem[\protect\citeauthoryear{Liu and Hauskrecht}{2016}]{liu2016learning}
Zitao Liu and Milos Hauskrecht.
\newblock Learning adaptive forecasting models from irregularly sampled
  multivariate clinical data.
\newblock In {\em Thirtieth AAAI Conference on Artificial Intelligence}, 2016.

\bibitem[\protect\citeauthoryear{Liu \bgroup \em et al.\egroup
  }{2013}]{liu2013modeling}
Zitao Liu, Lei Wu, and Milos Hauskrecht.
\newblock Modeling clinical time series using gaussian process sequences.
\newblock In {\em SIAM International Conference on Data Mining (SDM)}, pages
  623--631. SIAM, 2013.

\bibitem[\protect\citeauthoryear{Nguyen \bgroup \em et al.\egroup
  }{2017}]{nguyen2016deepr}
Phuoc Nguyen, Truyen Tran, Nilmini Wickramasinghe, and Svetha Venkatesh.
\newblock {Deepr: A Convolutional Net for Medical Records}.
\newblock {\em Journal of Biomedical and Health Informatics}, 21(1), 2017.

\bibitem[\protect\citeauthoryear{Pham \bgroup \em et al.\egroup
  }{2017}]{pham2017predicting}
Trang Pham, Truyen Tran, Dinh Phung, and Svetha Venkatesh.
\newblock Predicting healthcare trajectories from medical records: A deep
  learning approach.
\newblock {\em Journal of Biomedical Informatics}, 69:218--229, May 2017.

\bibitem[\protect\citeauthoryear{Razavian \bgroup \em et al.\egroup
  }{2016}]{razavian2016multi}
Narges Razavian, Jake Marcus, and David Sontag.
\newblock Multi-task prediction of disease onsets from longitudinal laboratory
  tests.
\newblock In {\em Proceedings of the 1st Machine Learning for Healthcare
  Conference}, pages 73--100, 2016.

\bibitem[\protect\citeauthoryear{Schulam and
  Saria}{2016}]{schulam2016integrative}
Peter Schulam and Suchi Saria.
\newblock Integrative analysis using coupled latent variable models for
  individualizing prognoses.
\newblock {\em Journal of Machine Learning Research}, 17:1--35, 2016.

\bibitem[\protect\citeauthoryear{Silva \bgroup \em et al.\egroup
  }{2012}]{silva2012predicting}
Ikaro Silva, George Moody, Daniel~J Scott, Leo~A Celi, and Roger~G Mark.
\newblock Predicting in-hospital mortality of icu patients: The
  physionet/computing in cardiology challenge 2012.
\newblock In {\em Computing in Cardiology (CinC), 2012}, pages 245--248. IEEE,
  2012.

\bibitem[\protect\citeauthoryear{Tresp and Briegel}{1998}]{tresp1998solution}
Volker Tresp and Thomas Briegel.
\newblock A solution for missing data in recurrent neural networks with an
  application to blood glucose prediction.
\newblock {\em Advances in Neural Information Processing Systems}, pages
  971--977, 1998.

\end{thebibliography}

\end{document}